# $\mathcal{TDL}$—A Type Description Language for Constraint-Based Grammars

Hans-Ulrich Krieger, Ulrich Schäfer
{krieger,schaefer}@dfki.uni-sb.de
German Research Center for Artificial Intelligence (DFKI)
Stuhlsatzenhausweg 3, D-66123 Saarbrücken, Germany

## Abstract

This paper presents $\mathcal{TDL}$, a typed feature-based representation language and inference system. Type definitions in $\mathcal{TDL}$ consist of type and feature constraints over the boolean connectives. $\mathcal{TDL}$ supports open- and closed-world reasoning over types and allows for partitions and incompatible types. Working with partially as well as with fully expanded types is possible. Efficient reasoning in $\mathcal{TDL}$ is accomplished through specialized modules.
**Topical Paper. Topic Area:** software for NLP, grammar formalism for typed feature structures.

## 1 Introduction

Over the last few years, constraint-based grammar formalisms have become the predominant paradigm in natural language processing and computational linguistics. Their success stems from the fact that they can be seen as a monotonic, high-level representation language for linguistic knowledge which can be given a precise mathematical semantics. The main idea of representing as much linguistic knowledge as possible through a unique data type called *feature structure*, allows the integration of different description levels without taking care of interface problems. While the first approaches relied on annotated phrase structure rules (e.g., PATR-II), modern formalisms try to specify grammatical knowledge as well as lexicon entries entirely through feature structures. In order to achieve this goal, one must enrich the expressive power of the first unification-based formalisms with different forms of disjunctive descriptions. Later, other operations came into play, e.g., (classical) negation. Other proposals consider the integration of functional/relational dependencies into the formalism which make them in general Turing-complete (e.g., ALE [4]). However the most important extension to formalisms consists of the incorporation of *types*, for instance in modern systems like TFS [15], CUF [6], or $\mathcal{TDL}$ [7]. Types are ordered hierarchically as it is known from object-oriented programming languages. This leads to multiple inheritance in the description of linguistic entities. Finally, recursive types are necessary to describe at least phrase-structure recursion which is inherent in all grammar formalisms which are not provided with a context-free backbone.

In the next section, we argue for the need and relevance of using types in CL and NLP. After that, we give an overview of $\mathcal{TDL}$ and its specialized inference modules. Especially, we have a closer look on the novel features of $\mathcal{TDL}$ and present the techniques we have employed in implementing $\mathcal{TDL}$.

## 2 Motivation

Modern typed unification-based grammar formalisms differ from early untyped systems in that they highlight the notion of a *feature type*. Types can be arranged hierarchically, where a subtype *inherits* monotonically all the information from its supertypes and unification plays the role of the primary information-combining operation. A *type definition* can be seen as an abbreviation for a complex expression, consisting of type constraints (concerning the sub-/supertype relationship) and feature constraints (stating the appropriate attributes and their values) over the connectives $\land$, $\lor$, and $\neg$. Types serve as abbreviations for lexicon entries, ID rule schemata, and universal as well as language-specific principles as is familiar from HPSG. Besides using types as an abbreviational means as templates are, there are other advantages as well which cannot be accomplished by templates:

- STRUCTURING KNOWLEDGE
  Types together with the possibility to order them hierarchically allow for a modular and clean way to represent linguistic knowledge adequately. Moreover, generalizations can be put at the appropriate levels of representation.

- EFFICIENT PROCESSING
  Certain type constraints can be compiled into efficient representations like bit vectors [1], where a GLB (greatest lower bound), LUB (least upper bound), or a $\preceq$ (type subsumption) computation reduces to low-level bit manipulation; see Section 3.2. Moreover, types release untyped unification from expensive computation through the possibility to declare them incompatible. In addition, working with type names only or with partially expanded types minimizes the costs of copying structures during processing. This can only be accomplished if the system makes a mechanism for type expansion available; see Section 3.4.

- TYPE CHECKING
  Type definitions allow a grammarian to declare which attributes are appropriate for a given type and which types are appropriate for a given attribute, therefore disallowing one to write inconsistent feature structures. Again, type expansion is necessary to determine the global consistency of a given description.

- RECURSIVE TYPES
  Recursive types give a grammar writer the opportunity to formulate certain functions or relations as recursive type specifications. Working in the type deduction paradigm enforces a grammar writer to replace the context-free back-bone through recursive types. Here, parameterized delayed type expansion is the ticket to the world of controlled linguistic deduction [13]; see Section 3.4.

## 3  *TDL*

*TDL* is a unification-based grammar development environment and run time system supporting HPSG-like grammars. Work on *TDL* has started within the DISCO project of the DFKI [14] (this volume). The DISCO grammar currently consists of approx. 900 type specifications written in *TDL* and is the largest HPSG grammar for German [9]. The core engine of DISCO consists of *TDL* and the feature constraint solver *UDiNe* [3]. *UDiNe* itself is a powerful untyped unification machinery which allows the use of distributed disjunctions, general negation, and functional dependencies. The modules communicate through an interface, and this connection mirrors exactly the way an abstract typed unification algorithm works: two typed feature structures can only be unified if the attached types are definitely compatible. This is accomplished by the unifier in that *UDiNe* handles over two typed feature structures to *TDL* which gives back a simplified form (plus additional information; see Fig. 1). The motivation for separating type and feature constraints and processing them in specialized modules (which again might consist of specialized components as is the case in *TDL*) is twofold: (i) this strategy reduces the complexity of the whole system, thus making the architecture clear, and (ii) leads to a higher performance of the whole system because every module is designed to cover only a specialized task.

### 3.1  *TDL* Language

*TDL* supports type definitions consisting of type constraints and feature constraints over the operators $\wedge$, $\vee$, $\neg$, and $\oplus$ (xor). The operators are generalized in that they can connect feature descriptions, coreference tags (logical variables) as well as types. *TDL* distinguishes between *avm types* (open-world semantics), *sort types* (closed-world semantics), *built-in types* (being made available by the underlying COMMON LISP system), and *atoms*. Recursive types are explicitly allowed and handled by a sophisticated lazy type expansion mechanism.

In asking for the greatest lower bound of two avm types $a$ and $b$ which share no common subtype, *TDL* always returns $a \wedge b$ (open-world reasoning), and not $\bot$. The reason for assuming this is manifold: (i) partiality of our linguistic knowledge, (ii) approach is in harmony with terminological (KL-ONE-like) languages which share a similar semantics, (iii) important during incremental grammar/lexicon construction (which has been shown useful in our project), and (iv) one must not write superfluous type definitions to guarantee successful type unifications during processing.

The opposite case holds for the GLB of sort types (closed-world approach). Furthermore, sort types differ in another point from avm types in that they are not further structured, as is the case for atoms. Moreover, *TDL* offers the possibility to declare *partitions*, a feature heavily used in HPSG. In addition, one can declare sets of types as *incompatible*, meaning that the conjunction of them yields $\bot$, so that specific avm types can be closed.

*TDL* allows a grammarian to define and use *parameterized templates* (macros). There exists a special *instance definition facility* to ease the writing of lexicon entries which differ from normal types in that they are not entered into the type hierarchy. Input given to *TDL* is parsed by a Zebu-generated LALR(1) parser [8] to allow for an intuitive, *high-level input syntax* and to abstract from uninteresting details imposed by the unifier and the underlying LISP system.

The kernel of *TDL* (and of most other monotonic systems) can be given a set-theoretical semantics along the lines of [12]. It is easy to translate *TDL* statements into denotation-preserving expressions of Smolka's feature logic, thus viewing *TDL* only as syntactic sugar for a restricted (decidable) subset of first-order logic. Take for instance the following feature description $\phi$ written as an attribute-value matrix:

$$\phi = \begin{bmatrix} np \\ \text{AGR} \;\; \boxed{x} \;\; \begin{bmatrix} agreement \\ \text{NUM} \; sg \\ \text{PERS} \; 3rd \end{bmatrix} \\ \text{SUBJ} \;\; \boxed{x} \end{bmatrix}$$

It is not hard to rewrite this two-dimensional description to a flat first-order formula, where attributes/features (e.g., AGR) are interpreted as binary relations and types (e.g., $np$) as unary predicates:

$$\exists x \,.\, np(\phi) \wedge \text{AGR}(\phi, x) \wedge agreement(x) \wedge \\ \text{NUM}(x, sg) \wedge \text{PERS}(x, 3rd) \wedge \text{SUBJ}(\phi, x)$$

The corresponding *TDL* type definition of $\phi$ looks as follows (actually & is used on the keyboard instead of $\wedge$, | instead of $\vee$, ˜ instead of $\neg$):

$$\phi := np \wedge [\text{AGR} \; \#x \wedge agreement \wedge [\text{NUM} \; sg, \text{PERS} \; 3rd], \\ \text{SUBJ} \; \#x].$$

### 3.2  Type Hierarchy

The type hierarchy is either called directly by the control machinery of *TDL* during the definition of a type (type classification) or indirectly via the simplifier both at definition and at run time (type unification).

#### 3.2.1  Encoding Method

The implementation of the type hierarchy is based on Aït-Kaci's encoding technique for partial orders [1]. Every type $t$ is assigned a code $\gamma(t)$ (represented via a bit vector) such that $\gamma(t)$ reflects the reflexive transitive closure of the subsumption relation with respect to $t$. Decoding a code $c$ is realized either by a look-up (iff $\exists t \,.\, \gamma^{-1}(c) = t$) or by computing the "maximal restriction" of the set of types whose codes are less than $c$. Depending on the encoding method, the hierarchy occupies $O(n \log n)$ (compact encoding) resp. $O(n^2)$ (transitive closure encoding) bits. Here, GLB/LUB operations directly correspond to bit-or/and instructions. GLB, LUB and $\preceq$ computations have the nice property that they can be carried out in this framework in $O(n)$, where $n$ is the

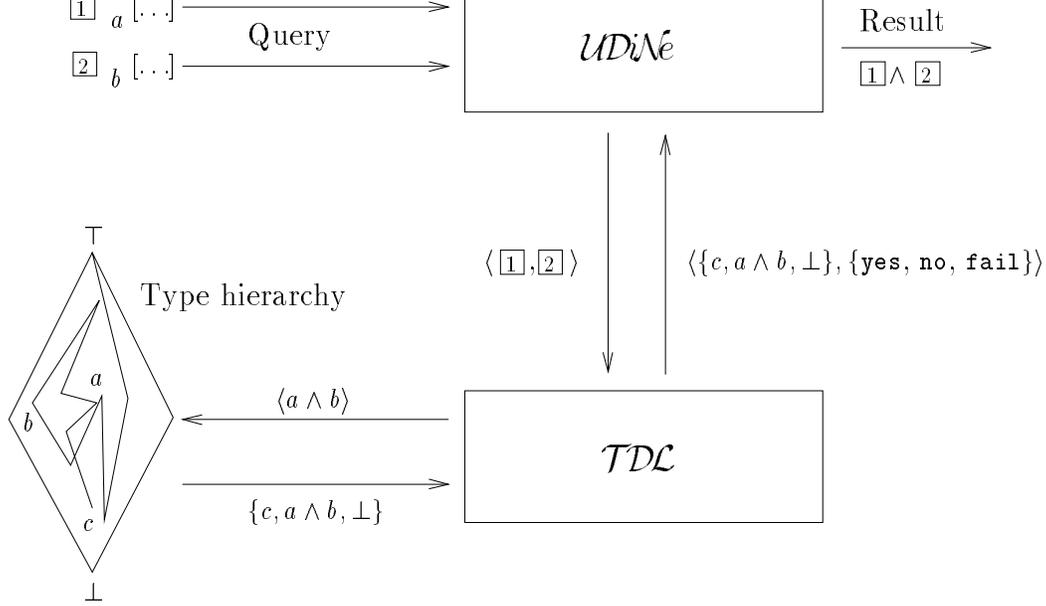

Figure 1: **Interface between $\mathcal{TDL}$ and $\mathcal{UDiNe}$.** Depending on the type hierarchy and the type of [1] and [2], $\mathcal{TDL}$ either returns c (c is definitely the GLB of a and b) or $a \wedge b$ (open-world reasoning) resp. $\bot$ (closed-world reasoning) if there doesn't exist a single type which is equal to the GLB of a and b. In addition, $\mathcal{TDL}$ determines whether $\mathcal{UDiNe}$ must carry out feature term unification (**yes**) or not (**no**), i.e., the return type contains all the information one needs to work on properly (**fail** signals a global unification failure).

number of types.[1]

Aït-Kaci's method has been extended in $\mathcal{TDL}$ to cover the open-world nature of avm types in that potential GLB/LUB candidates (calculated from their codes) must be verified. Why so? Take the following example to see why this is necessary:

$$x := y \wedge z$$
$$x' := y' \wedge z' \wedge [a\ 1]$$

During processing, one can definitely substitute $y \wedge z$ through $x$, but rewriting $y' \wedge z'$ to $x'$ is not correct, because $x'$ differs from $y' \wedge z'$—$x'$ is more specific as a consequence of the feature constraint $[a\ 1]$. So we make a distinction between the "internal" greatest lower bound $\text{GLB}_{\preceq}$, concerning only the type subsumption relation by using Aït-Kaci's method alone (which is however used for sort types) and the "external" one, $\text{GLB}_{\sqsubseteq}$, which takes the subsumption relation over feature structures into account.

With $\text{GLB}_{\preceq}$ and $\text{GLB}_{\sqsubseteq}$ in mind, we can define a generalized GLB operation informally by the following table. This GLB operation is actually used during type unification ($fc$ = feature constraint):

| GLB | $avm_1$ | $sort_1$ | $atom_1$ | $fc_1$ |
|---|---|---|---|---|
| $avm_2$ | see 1. | $\bot$ | $\bot$ | see 2. |
| $sort_2$ | $\bot$ | see 3. | see 4. | $\bot$ |
| $atom_2$ | $\bot$ | see 4. | see 5. | $\bot$ |
| $fc_2$ | see 2. | $\bot$ | $\bot$ | see 6. |

*where*

---

[1] Actually, one can choose in $\mathcal{TDL}$ between the two encoding techniques and between bit vectors and bignums in COMMON LISP for the representation of the codes. In our LISP implementation, operations on bignums are a magnitude faster than on bit vectors.

1. $\begin{cases} avm_3 \iff \text{GLB}_{\sqsubseteq}(avm_1, avm_2) = avm_3 \\ avm_1 \iff avm_1 = avm_2 \\ \bot \iff \text{GLB}_{\preceq}(avm_1, avm_2) = \bot, \text{ via an} \\ \qquad \text{explicit incompatibility declaration} \\ avm_1 \wedge avm_2, \text{ otherwise (open world)} \end{cases}$

2. $\begin{cases} avm_{1,2} \iff \text{expand}(avm_{1,2}) \sqcap fc_{2,1} \neq \bot \\ \bot, \text{ otherwise} \end{cases}$

3. $\begin{cases} sort_3 \iff \text{GLB}_{\preceq}(sort_1, sort_2) = sort_3 \\ sort_1 \iff sort_1 = sort_2 \\ \bot, \text{ otherwise (closed world)} \end{cases}$

4. $\begin{cases} atom_{1,2} \iff \text{type-of}(atom_{1,2}) \preceq sort_{2,1}, \\ \qquad \text{where } sort_{2,1} \text{ is a built-in} \\ \bot, \text{ otherwise} \end{cases}$

5. $\begin{cases} atom_1 \iff atom_1 = atom_2 \\ \bot, \text{ otherwise} \end{cases}$

6. $\begin{cases} \top \iff fc_1 \sqcap fc_2 \neq \bot \\ \bot, \text{ otherwise} \end{cases}$

The encoding algorithm is also extended towards the *redefinition of types* and the *use of undefined types*, an essential part of an incremental grammar/lexicon development system. Redefining a type means not only to make changes local to this type. Instead, one has to redefine all *dependents* of this type—all subtypes in case of a conjunctive type definition and all disjunction alternatives for a disjunctive type specification plus, in both cases, all types which use these types in their definition. The dependent types of a type $t$ can be characterized graph-theoretically via the strongly connected component of $t$ with respect to the dependency relation.

### 3.2.2 Decomposing Type Definitions

*Conjunctive*, e.g., $x := y \wedge z$ and *disjunctive type specifications*, e.g., $x' := y' \vee z'$ are entered differently into the hierarchy: $x$ inherits from its supertypes $y$ and $z$, whereas $x'$ defines itself through its alternatives $y'$ and $z'$. This distinction is represented through the use of different kinds of edges in the type graph (bold edges denote disjunction elements; see Fig. 3). But it is worth noting that both of them express *subsumption* ($x \preceq y$ and $x' \succeq y'$) and that the GLB/LUB operations must work properly over "conjunctive" as well as "disjunctive" subsumption links.

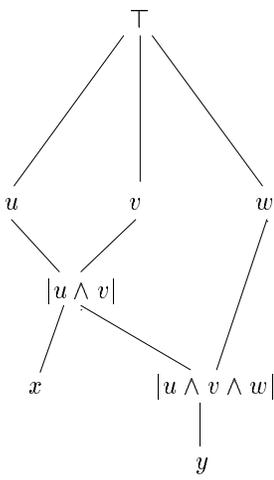
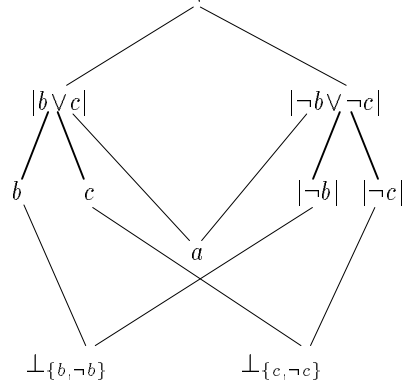

Figure 2: *The intermediate types $|u \wedge v|$ and $|u \wedge v \wedge w|$ are introduced by $\mathcal{TDL}$ during the type definitions $x := u \wedge v \wedge [a\ 0]$ and $y := w \wedge v \wedge u \wedge [a\ 1]$.*

$\mathcal{TDL}$ *decomposes* complex definitions consisting of $\wedge$, $\vee$, and $\neg$ by introducing *intermediate types*, so that the resulting expression is either a *pure conjunction or a disjunction of type symbols*. Intermediate type names are enclosed in vertical bars (cf. the intermediate types $|u \wedge v|$ and $|u \wedge v \wedge w|$ in Fig. 2).

The same technique is applied when using $\oplus$ (see Fig. 3). $\oplus$ will be decomposed into $\wedge$, $\vee$ and $\neg$, plus additional intermediates. For each negated type $\neg t$, $\mathcal{TDL}$ introduces a new intermediate type symbol $|\neg t|$ having the definition $\neg t$ and declares it incompatible with $t$ (see Section 3.2.3). In addition, if $t$ is not already present, $\mathcal{TDL}$ will add $t$ as a new type to the hierarchy (see types $|\neg b|$ and $|\neg c|$ in Fig. 3).

Let's consider the example $a := b \oplus c$. The decomposition can be stated informally by the following rewrite steps (assuming that the user has chosen CNF):

$$\frac{a := b \oplus c}{\frac{a := (b \wedge \neg c) \vee (\neg b \wedge c)}{\frac{a := (b \vee \neg b) \wedge (b \vee c) \wedge (\neg b \vee \neg c) \wedge (\neg c \vee c)}{\frac{a := (b \vee c) \wedge (\neg b \vee \neg c)}{a := |b \vee c| \wedge |\neg b \vee \neg c|}}}}$$

### 3.2.3 Incompatible Types and Bottom Propagation

*Incompatible types* lead to the introduction of specialized bottom symbols (see Fig. 3 and 4) which however are identified in the underlying logic in that they denote the empty set. These bottom symbols must be propagated downwards by a mechanism called *bottom propagation* which takes place at definition time (see Fig. 4). Note that it is important to take not only subtypes of incompatible types into account but also disjunction elements as the following example shows:

Figure 3: *Decomposing $a := b \oplus c$, such that $a$ inherits from the intermediates $|b \vee c|$ and $|\neg b \vee \neg c|$.*

$$\left.\begin{array}{l}\bot = a \wedge b. \\ b := b_1 \vee b_2.\end{array}\right\} \xrightarrow{BP} a \wedge b_1 = \bot \text{ and } a \wedge b_2 = \bot$$

One might expect that incompatibility statements together with feature term unification no longer lead to a monotonic, set-theoretical semantics. But this is not the case. To preserve monotonicity, one must assume a *2-level interpretation of typed feature structures*, where feature constraints and type constraints might denote different sets of objects and the global interpretation is determined by the intersection of the two sets. Take for instance the type definitions $A := [a\ 1]$ and $B := [b\ 1]$, plus the user declaration $\bot = A \wedge B$, meaning that $A$ and $B$ are incompatible. Then $A \wedge B$ will simplify to $\bot$ although the corresponding feature structures of $A$ and $B$ successfully unify to $[a\ 1, b\ 1]$, thus the global interpretation is $\bot$.

### 3.3 Symbolic Simplifier

The simplifier operates on arbitrary $\mathcal{TDL}$ expressions. Simplification is done at definition time and at run time when typed unification takes place (cf. Fig. 1). The main issue of symbolic simplification is to avoid (i) unnecessary feature constraint unification and (ii) queries to the type hierarchy by simply applying "syntactic" reduction rules. Consider an expression like $x_1 \wedge \cdots \wedge x_i \cdots \wedge \neg x_i \ldots \wedge x_n$. The simplifier will detect $\bot$ by simply applying reduction rules.

The simplification schemata are well known from the propositional calculus. They are hard-wired in the implementation to speed up computation. Formally, type simplification in $\mathcal{TDL}$ can be characterized as a term rewriting system. A set of reduction rules is applied until a *normal form* is reached. Confluence and termination is guaranteed by imposing a *total generalized lexicographic order* on terms (see below). In addition, this order has the nice effects of neglecting commutativity (which is expensive and might lead to termination problems): there is only one representative for a given formula. Therefore, *memoization* is cheap and is employed in $\mathcal{TDL}$ to reuse precomputed results of simplified expressions (one must not cover all permutations of a formula). Additional reduction rules are applied at run time using "semantic" information of the type hierarchy (GLB, LUB, and $\preceq$).

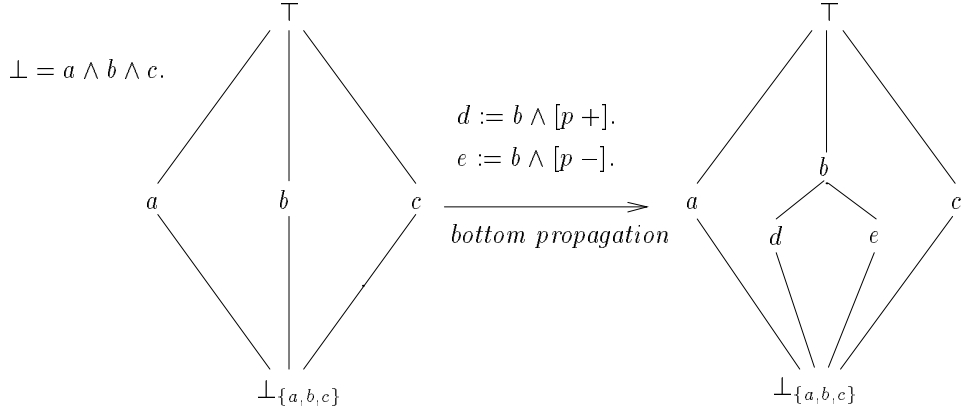

Figure 4: *Bottom propagation triggered through the subtypes d and e of b, so that $a \wedge d \wedge c$ as well as $a \wedge e \wedge c$ will simplify to $\bot$ during processing.*

### 3.3.1 Normal Form

In order to reduce an arbitrary type expression to a simpler expression, simplification rules must be applied. So we have to define what it means for an expression to be "simple". One can either choose the conjunctive or disjunctive normal form. The advantages of CNF/DNF are:

- UNIQUENESS
  Type expressions in normal form are unique modulo commutativity. Sorting type expressions according to a total lexicographic order will lead to a total uniqueness of type expressions (see Section 3.3.3).

- LINEARITY
  Type expressions in normal form are linear. Arbitrary nested expressions can be transformed into flat expressions. This may reduce the complexity of later simplifications, e.g., at run time.

- COMPARABILITY
  This property is a consequence of the two other properties. Unique and linear expressions make it easy to find or to compare (sub)expressions. This is important for the memoization technique described in Section 3.3.4.

### 3.3.2 Reduction Rules

In order to reach a normal form, it would suffice to apply only the schemata for double negation, distributivity, and De Morgan's laws. However, in the worst case, these three rules would blow up the length of the normal form to exponential size (compared with the number of literals in the original expression). To avoid this, other rules are used intermediately: idempotence, identity, absorption, etc. If they can be applied, they always reduce the length of the expressions. Especially at run time, but also at definition time, it is useful to exploit information from the type hierarchy. Further simplifications are possible by asking for the GLB, LUB, and $\preceq$.

### 3.3.3 Lexicographic Order

To avoid the application of the commutativity rule, we introduce a total lexicographic order on type expressions. Together with DNF/CNF, we obtain a unique sorted normal form for an arbitrary type expression. This guarantees fast comparability.

We define the order $<_{NF}$ on n-ary normal forms: *type $<_{NF}$ negated type $<_{NF}$ conjunction $<_{NF}$ disjunction $<_{NF}$ symbol $<_{NF}$ string $<_{NF}$ number*. For the comparison of atoms, strings, and type names, we use the lexicographical order on strings and for numbers the ordering $<$ on natural numbers.

Example: $a <_{NF} b <_{NF} bb <_{NF} \neg a <_{NF} a \wedge b <_{NF} a \wedge \neg a <_{NF} a \vee b <_{NF} a \vee b \vee c <_{NF} a \vee 1$

### 3.3.4 Memoization

The memoization technique described in [10] has been adapted in order to reuse precomputed results of type simplification. The lexicographically sorted normal form guarantees fast access to precomputed type simplifications. Memoization results are also used by the recursive simplification algorithm to exploit precomputed results for subexpressions.

Some empirical results show the usefulness of memoization. The current DISCO grammar for German consists of 885 types and 27 templates. After a full type expansion of a toy lexicon of 244 instances/entries, the memoization table contains approx. 3000 entries (literals are not memoized). 18000 results have been reused at least once (some up to 600 times) of which 90 % are proper simplifications (i.e., the simplified formulae are really shorter than the unsimplified ones).

## 3.4 Type Expansion and Control

We noted earlier that types allow us to refer to complex constraints through the use of symbol names. Reconstructing the constraints which determine a type (represented as a feature structure) requires a complex operation called *type expansion*. This is comparable to Carpenter's *totally well-typedness* [5].

### 3.4.1 Motivation

In $\mathcal{TDL}$, the motivation for type expansion is manifold:

- CONSISTENCY
  At definition time, type expansion determines whether the set of type definitions (grammar and lexicon) is consistent. At run time, type expansion is involved in checking the satisfiability of the unification of two partially expanded typed feature structures, e.g., during parsing.

- ECONOMY
  From the standpoint of efficiency, it does make sense to work only with small, partially expanded structures (if possible) to speed up feature term unification and to reduce the amount of copying. At the end of processing however, one has to make the result/constraints explicit.
- RECURSION
  Recursive types are inherently present in modern constraint-based grammar theories like HPSG which are not provided with a context-free backbone. Moreover, if the formalism does not allow functional or relational constraints, one must specify certain functions/relations like *append* through recursive types. Take for instance Aït-Kaci's version of the *append* type which can be stated in $\mathcal{TDL}$ as follows:

$$append := append_0 \vee append_1.$$
$$append_0 := [\text{FRONT} < >,$$
$$\qquad \text{BACK} \#1 \wedge list,$$
$$\qquad \text{WHOLE} \#1].$$
$$append_1 := [\text{FRONT} < \#first \cdot \#rest1 >,$$
$$\qquad \text{BACK} \#back \wedge list,$$
$$\qquad \text{WHOLE} < \#first \cdot \#rest2 >,$$
$$\qquad \text{PATCH } append \wedge [\text{FRONT} \#rest1,$$
$$\qquad\qquad \text{BACK} \#back,$$
$$\qquad\qquad \text{WHOLE} \#rest2]].$$

- TYPE DEDUCTION
  Parsing and generation can be seen in the light of type deduction as a uniform process, where ideally only the phonology (for parsing) or the semantics (for generation) must be given. Type expansion together with a sufficiently specified grammar then is responsible in both cases for constructing a fully specified feature structure which is maximal informative and compatible with the input. However, [15] has shown that type expansion without sophisticated control strategies is in many cases inefficient and moreover does not guarantee termination.

### 3.4.2 Controlled Type Expansion

Uszkoreit [13] introduced a new strategy for linguistic processing called *controlled linguistic deduction*. His approach permits the specification of linguistic performance models without giving up the declarative basis of linguistic competence, especially monotonicity and completeness. The evaluation of both conjunctive and disjunctive constraints can be *controlled* in this framework. For conjunctive constraints, the one with the highest failure probability should be evaluated first. For disjunctive ones, a success measure is used instead: the alternative with the highest success probability is used until a unification fails, in which case one has to backtrack to the next best alternative.

$\mathcal{TDL}$ and $\mathcal{UDiNe}$ support this strategy in that every feature structure can be associated with its success/failure potential such that type expansion can be sensitive to these settings. Moreover, one can make other decisions as well during type expansion:

- only regard structures which are subsumed by a given type resp. the opposite case (e.g., expand the type *subcat-list* always or never expand the type *daughters*)
- take into account only structures under certain paths or again assume the opposite case (e.g., always expand the value under path SYNSEM|LOC|CAT; in addition, it is possible to employ path patterns in the sense of pattern matching)
- set the depth of type expansion for a given type

Note that we are not restricted to apply only one of these settings—they can be used in combination and can be changed dynamically during processing. It does make sense, for instance, to expand at certain well-defined points during parsing the (partial) information obtained so far. If this will not result in a failure, one can throw away (resp. store) this fully expanded feature structure, working on with the older (and smaller) one. However, if the information is inconsistent, we must backtrack to older stages in computation. Going this way which of course assumes heuristic knowledge (language as well as grammar-specific knowledge) results in faster processing and copying. Moreover, the inference engine must be able to handle possibly inconsistent knowledge, e.g., in case of a chart parser to allow for a third kind of edge (besides active and passive ones).

### 3.4.3 Recursive Types, Implementational Issues, and Undecidability

The set of all recursive types of a given grammar/lexicon can be precompiled by employing the dependency graph of this type system. This graph is updated every time a new type definition is added to the system. Thus detecting whether a given type is recursive or not reduces to a simple table look-up. However the expansion of a recursive type itself is a little bit harder. In $\mathcal{TDL}$, we are using a *lazy* expansion technique which only makes those constraints explicit which are really new. To put it in another way: if no (global or local) control information is specified to guide a specific expansion, a recursive type will be be expanded under all its paths (local plus inherited paths) until one reaches a point where the information is already given in a *prefix* path. We call such an expanded structure a *resolved* typed feature structure. Of course, there are infinitely many resolved feature structures, but this structure is the *most general* resolved one.

Take for instance the *append* example from the previous section. *append* is of course a recursive type because one of its alternatives, viz., $append_1$ uses *append* under the PATCH attribute. Expanding *append* with no additional information supplied (especially no path leading inside $append_1$, e.g., PATCH|PATCH|PATCH) yields a disjunctive feature structure where both $append_0$ and $append_1$ are substituted by their definition. The expansion then stops if no other information enforce a further expansion.

In practice, one has to keep track of the visited paths and visited typed feature structures to avoid unnecessary expansion. To make expansion more efficient, we mark structures whether they are fully expanded or not. A feature structure is then fully expanded iff all of its substructures are fully expanded. This simple idea leads to a massive reduction of the search space when dealing with incremental expansion (e.g., during parsing).

It is worth noting that the satisfiability of feature descriptions admitting recursive type equations/definitions is in general undecidable. Rounds and Manaster-Ramer [11] were the first having shown that a Kasper-Rounds logic enriched with recursive types allows one to encode a Turing machine. Because our logic is much more richer, we immediately get the same result for $\mathcal{TDL}$.

However, one can choose in $\mathcal{TDL}$ between a complete expansion algorithm which may not terminate and a non-complete one to guarantee termination (see [2] and [5, Ch. 15] for similar proposals). The latter case heavily depends on the notion of resolvedness (see above). In both cases, the depth of the search space can be restricted by specifying a maximal path length.

## 4  Comparison with other Systems

$\mathcal{TDL}$ is unique in that it implements many novel features not found in other systems like ALE [4], LIFE [2], or TFS [15]. Of course, these systems provide other features which are not present in our formalism. What makes $\mathcal{TDL}$ unique in comparison to them is the distinction open vs. closed world, the availability of the full boolean connectives and distributed disjunctions (via $\mathcal{UDiNe}$), as well as an implemented lazy type expansion mechanism for recursive types (as compared with LIFE). ALE, for instance, neither allows disjunctive nor recursive types and enforces the type hierarchy to be a BCPO. However, it makes recursion available through definite relations and incorporates special mechanisms for empty categories and lexical rules. TFS comes up with a closed world, the unavailability of negative information (only implicitly present) and only a poor form of disjunctive information but performs parsing and generation entirely through type deduction (in fact, it was the first system). LIFE comes closest to us but provides a semantics for types that is similar to TFS. Moreover the lack of negative information and distributed disjunctions makes it again comparable with TFS. LIFE as a whole can be seen as an extension of PROLOG (as was the case for its predecessor LOGIN), where first-order terms are replaced by $\psi$-terms. In this sense, LIFE is richer than our fomalism in that it offers a full relational calculus.

## 5  Summary and Outlook

In this paper, we have presented $\mathcal{TDL}$, a typed feature formalism that integrates a powerful feature constraint solver and type system. Both of them provide the boolean connectives $\wedge$, $\vee$, and $\neg$, where a complex expression is decomposed by employing intermediate types. Moreover, recursive types are supported as well. In $\mathcal{TDL}$, a grammar writer decides whether types live in an open or a closed world. This effects GLB and LUB computations. The type system itself consists of several inference components, each designed to cover efficiently a specific task: (i) a bit vector encoding of the hierarchy, (ii) a fast symbolic simplifier for complex type expressions, (iii) memoization to cache precomputed results, and (iv) a sophisticated type expansion mechanism. The system as described in this paper has been implemented in COMMON LISP and integrated in the DISCO environment [14].

The next major version of $\mathcal{TDL}$ will be integrated into a declarative specification language which allows linguists to define control knowledge that can be used during processing. In addition, certain forms of knowledge compilation will be made available in future versions of $\mathcal{TDL}$, e.g., the automatic detection of syntactic incompatibilities between types, so that a type computation can substitute an extensive feature term unification.